\documentclass[10pt,twocolumn,letterpaper]{article}

\usepackage{cvpr}
\usepackage{times}
\usepackage{epsfig}
\usepackage{graphicx}
\usepackage{amsmath}
\usepackage{amssymb}

\usepackage{subfigure}
\usepackage{color}
\usepackage{epstopdf}
\usepackage{booktabs}
\usepackage{multirow}
\usepackage{mathtools}
\usepackage{enumitem}

\usepackage[pagebackref=true,breaklinks=true,letterpaper=true,colorlinks,bookmarks=false]{hyperref}

\cvprfinalcopy 

\def\cvprPaperID{}

\ifcvprfinal\fi

\begin{document}

\title{MTLE: A Multitask Learning Encoder of Visual Feature Representations for Video and Movie Description }

\author{Oliver Nina\\
The Ohio State University\\
Columbus, OH\\
{\tt\small nina.3@osu.edu}
\and
Washington Garcia\\
University of Florida\\
Gainsville, FL\\
{\tt\small w.garcia@ufl.edu}
\and
Scott Clouse\\
Air Force Research Lab\\
Dayton, OH\\
{\tt\small hsclouse@ieee.org}
\and
Alper Yilmaz\\
The Ohio State University\\
Columbus, OH\\
{\tt\small yilmaz.15@osu.edu}
}

\maketitle

\begin{abstract}
Learning visual feature representations for video analysis is a daunting task that requires a large amount of training samples and a proper generalization framework.
Many of the current state of the art methods for video captioning and movie description rely on simple encoding mechanisms through recurrent neural networks to encode temporal visual information extracted from video data. In this paper, we introduce a novel multitask encoder-decoder framework for automatic semantic description and captioning of video sequences. In contrast to current approaches, our method relies on distinct decoders that train a visual encoder in a multitask fashion. Our system does not depend solely on multiple labels and allows for a lack of training data working even with datasets where only one single annotation is viable per video. Our method shows improved performance over current state of the art methods in several metrics on multi-caption and single-caption datasets. To the best of our knowledge, our method is the first method to use a multi-task approach for encoding video features. Our method demonstrates its robustness on the Large Scale Movie Description Challenge (LSMDC) 2017 where our method won the movie description task and its results were ranked among other competitors as the most helpful for the visually impaired. 

\end{abstract}

\begin{figure}
  \centering
  \includegraphics[width=.45\textwidth]{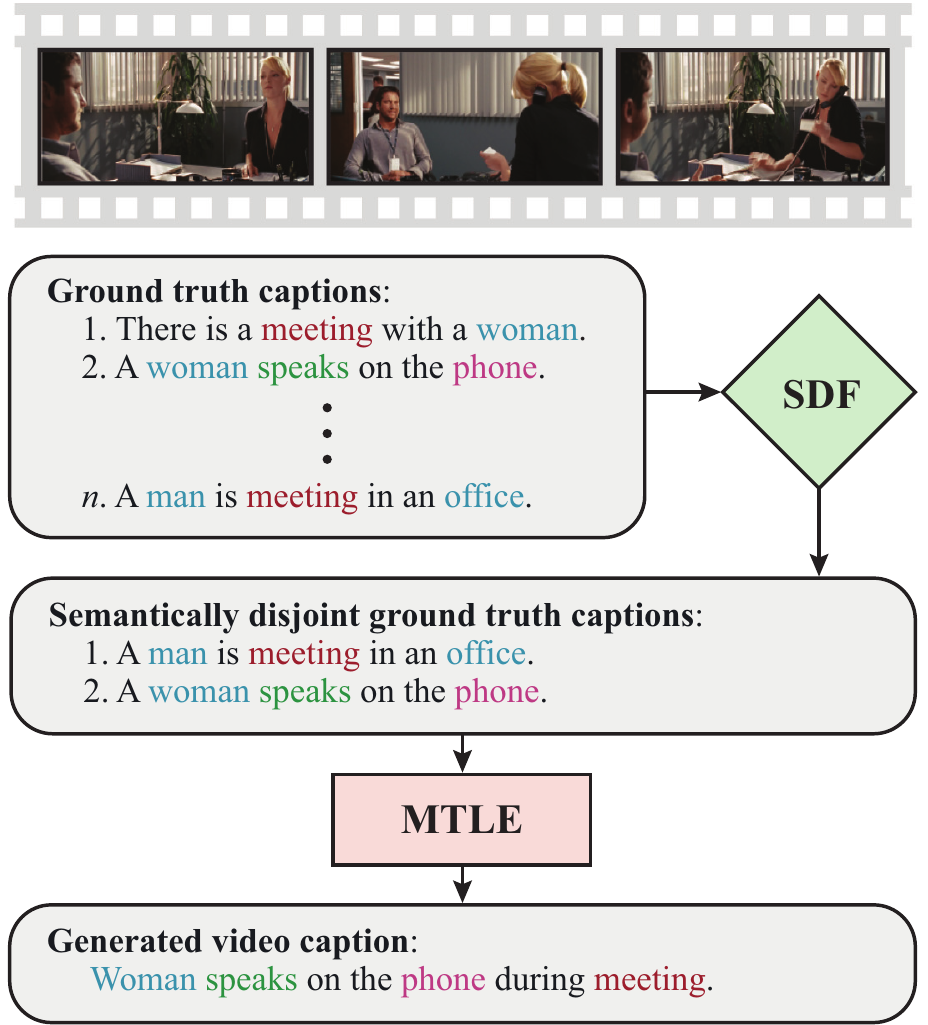}
  \vspace{0.5em}
  \caption{Example of MTLE-based video captioning. Ground truth captions are compared with a semantic distance function (SDF) to find semantically disjoint caption samples, which are passed to our MTLE method to produce a a video's caption. }
  \label{fig:sampleshow}
\end{figure}

\section{Introduction}

Video captioning and semantic video description has generated increasing attention in recent years due to encouraging results observed for similar problems such as  image captioning~\cite{vinyals2017show} and question and answering systems~\cite{antol2015vqa} where typical encoder-decoder frameworks are employed.  

Learning and encoding visual feature representations for video analysis and specifically video captioning is challenging. Some of these challenges come from the complexity and nature of the data where video frames displayed as images in a  time sequence add a temporal dimension to a much larger challenge of recognizing and detecting objects in a per frame basis. Because of the variable length of its temporal component, a paradigm that accommodates for an unconstrained sequence is necessary to process video frames. One way to accomplish this is through a feature encoding mechanism that is two-fold: 

First, visual CNN features are extracted from every frame using a CNN model previously trained on large single-image datasets such as ImageNet. The second step in the process is to transform or encode these CNN features into one single feature vector that is trained on video samples with a decoder conditioned on the encoder features and a cost function that relies on the provided labeled data in a supervised learning manner.  
 
Our method improves upon recent visual feature encoding methods by introducing a novel encoding framework consisted of an encoder which is jointly trained with multiple decoders in a multi-task fashion. Our encoder consists of a bi-directional recurrent neural network trained with multiple and separate decoders that use labeled samples to project the textual features into a semantic space where a distance metric can be used to compare semantic distances from different samples and captions.  

We use a semantic distance function that allows us to select captions being input to each decoder during training. Thus, given a set of different training labels for a single video/segment, we find the pair with the farthest semantic distance among all possible pairs. Intuitively, pairs that are farthest apart in the semantic space complement the meaning of each other while allowing us to obtain a centroid of a potential cluster in semantic space that will enable us to capture the full meaning of all the captions. This also avoids biasing the encoder weights towards a single sample, providing improved generalization. 

Our method is designed to work not only with multiple annotations per video but also single captions. Our objective function allows for a regularization term that will leverage multiple caption scenarios and augment training samples when there is only a limited number of training data. Thus, our method does not depend on a large number of training labels and can handle datasets with limited number of annotations such as LSMDC.

Our proposed method shows improvements over the current baseline in public datasets that contain multiple or single annotations per video such as MSVD~\cite{chen:acl11}, MSR-VTT~\cite{Xu2016}, TRECVID~\cite{trecvid} and LSMDC~\cite{lsmdc2015}. 

\begin{figure*}
  \centering
  \includegraphics[width=.95\textwidth]{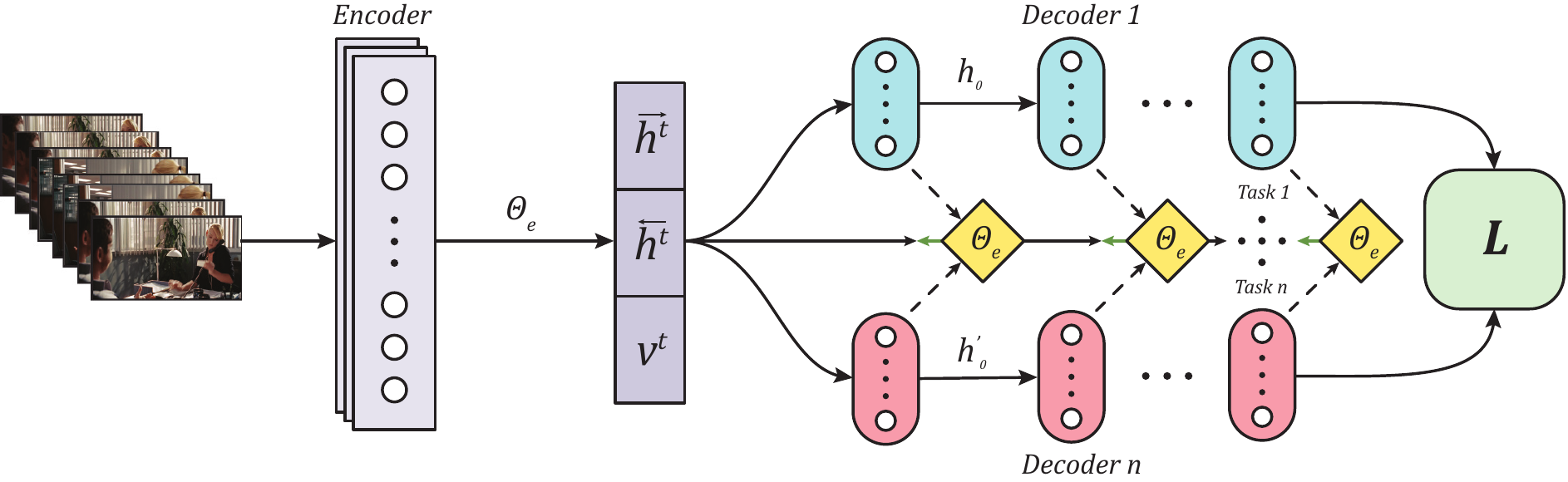}	
  \caption{Overview of our MTLE method. Video frames are passed to a CNN encoder for feature vector generation. From the encoder, feature vectors are passed to an RNN-based bi-directional attention encoder. The output of this process is the concatenated visual feature encoding of the video, which is passed to a multitask conditional decoder with soft attention. The parameters of the RNN-based bi-directional attention encoder are denoted by $\theta_e$, and are trained simultaneous to the multitask decoder. Green arrows denote back-propagation. }
  \label{fig:method}
\end{figure*}

\textbf{Contributions:} In essence, the contribution of this paper can be summarized in the following:
\begin{itemize}
\item We propose a novel type of visual encoder that uses multitask learning to improve generalization of encoder weights over large number of training samples. 
\item Using a multitask approach for learning visual representations is a non-linear problem which is difficult to solve through a convex optimization. Thus, we assume the tasks are linearly related and introduce a novel loss function for our multitask system that is used to train a new video encoder. 
\item Our framework allows us to have a limited number of training labels and captions for instance in single caption datasets where only one caption or description is provided per video. 
\end{itemize}
In this paper, we first give a brief introduction to the problem and discuss some issues with current methods and how our method helps overcome these challenges. We also give a brief literature review of past and current trends. We then explain our method in detail. We end with a discussion of our results and conclusion.
\section{Background}

Video captioning followed a similar evolution to image classification as traditional heuristic-based methods turned into deep learning frameworks. Early video captioning methods, for instance, evolved from low level image processing methods such as optical flow~\cite{kojima2002natural}, semantic event detection~\cite{lee2008save}, semantic content descriptors~\cite{Aradhye2009}, and object matching~\cite{Kollnig1994}.

Current methods for automatic video description use neural networks to model high-level representations of video based on individual frames~\cite{venugopalan2014translating,Yao2015,venugopalan15iccv}. These approaches use convolutional neural network (CNN) embeddings paired with a recurrent neural network (RNN) to form an encoder-decoder framework. The encoder creates frame-based video features from CNNs trained on image datasets. These features are then fed to a decoder to be ``decoded'' or translated into natural language. This framework automatically abstracts much of the low-level knowledge that was hand-crafted in early methods, allowing for more robust performance.

Different parts of the traditional encoder-decoder framework have been improved in recent years. Originally, in ~\cite{venugopalan2014translating}, the CNN's output features were averaged through a mean function and given as input to a stacked LSTM decoder. In~\cite{venugopalan15iccv}, video frame features were created using CNNs trained on RGB and optical flow videos, then combined as input to a single LSTM decoder. The downside of these previous implementations was their failure to capture naturally occurring temporal information from videos. In particular, they did not build a model that considered global and local temporal patterns. Yao et. al.~\cite{Yao2015} addressed this by introducing a novel attention mechanism whose weights are learned simultaneously with a single LSTM decoder. The attention mechanism is designed to exploit global temporal structure, while action features are used to encode local temporal structure. Action features are derived from a spatio-temporal convolutional network, and are used to augment frame appearance features. As in the previous methods, frame appearance features are obtained from a CNN model pre-trained on image datasets~\cite{szegedy2014going}. Although the work of Yao et. al. significantly improves the encoder-decoder framework with respect to temporal knowledge, it does not take advantage of important semantic information in ground truth data, which is the main focus of our work. 

Some methods present variations on RNN model structures, such as hierarchical recurrent neural networks, to amplify the knowledge gained from input frames. These include the work of Yu et al.~\cite{yu2016video} which uses a hierarchical neural network to augment the number of descriptions obtained from a single video. Instead of traditional stacked RNNs used for encoding of frame features,~\cite{pan2016hierarchical} uses a second LSTM layer on top of a regular LSTM feature encoder in order to reduce and sub-sample features from video frames. Nian et. al.~\cite{nian2017learning} aimed to find a mid-level representation of videos in the form of a spectrogram-inspired video response map (VRM), a single image that can represent important video attributes from a sequence of video frames. Although these methods encode better information from input frames, or encode the information in a new way, the information gain is not directly tied to annotator-generated knowledge, such as the semantic inferences an annotator makes during the ground truth labeling process.

Direct information gain can come from using representations of semantic concepts. The work in~\cite{pan2016video} introduces the use of Long Short Term Memory (LSTM) in combination with Transfer Semantic Attributes. The framework is trained jointly such that semantic attributes are complemented by image and video information, a step towards our goal of coupling semantic concepts with training data. Another step comes from ~\cite{SCN_CVPR2017}, which extended the normal LSTM used in previous systems ~\cite{venugopalan2014translating, temporal2014, Yao2015} by adding the detection of semantic concepts, or "tags". Tags are predicted by formulating the problem as a multi-label classification task, as in \cite{nian2017learning}. When decoding, these semantic-concept tags are used as weights to an ensemble of sets of LSTM parameters, whereas previous work only used tags during LSTM initialization~\cite{2015_lstm_image_Wu} or through soft attention~\cite{2016scn_attention}. This framework is used with video by running the semantic-concept prediction on representations of video. These systems offer a solid foundation for the inference of semantic concepts from video by introducing abstracted models of semantic information, such as "tags", into the learning problem. Although our goal is to infer semantic information, it is desirable to infer it from as close a human source as possible, as in the form of human-labeled video annotations, rather than a pre-trained model's prediction.

Our goal is thus to harvest the largest amount of semantic information possible from human-labeled video datasets. Although Zeng et. al.~\cite{Zeng2016} originally aimed to generate salient titles for raw, untrimmed user-generated videos, the authors also proposed the use of additional sentence-only examples to increase the amount of semantic information associated with the training data. The authors collect open-domain curated videos from online communities, and associate videos with their user-generated titles and descriptions, forming an ``in the wild" dataset. Ground truth titles are augmented by sampling additional titles from a very large YouTube title corpus. 

Our approach differs from current approaches and specifically traditional encoder-decoder methods, where we use a multitask learning approach to improve the learning and training of the encoder parameters that in consequence will improve the learning of the decoder parameters in a  symbiotic type of way. In the next sections, we explain in detail our method to show how it leverages other state of the art methods for semantic video description.

\section{Method}
In this section, we give an overview of a video captioning system and how it is extended to fit a multitask approach. This overview is summarized in Figure \ref{fig:method}.

Given a video $\mathbf{F}$ composed of $t$ frames $\mathbf{f}$ such that $\mathbf{F}=\{\mathbf{f}^1,..,\mathbf{f}^t\}$ and such is associated with a caption $\mathbf{X}=(\mathbf{x}_1,...,\mathbf{x}_i)$, where $\mathbf{x}$ is a one-hot vector of a vocabulary $V$ and $i$ is the size of the caption. We first process the video features and explain the encoding phase in order to reduce the number of temporal sequence features of a video.

\subsection{Visual Feature Encoding}
\label{sec:encoder}
Visual features from each video  are processed individually for every frame $\mathbf{f}$ separately to obtain $\mathbf{u}=\zeta(\mathbf{f})$ where $\zeta$ corresponds to the last layer of a CNN model. Because CNN features are specially large when deeper models are used such as ResNet, it is rather crucial to reduce such CNN features of size $D$ to $r$-dimensional features with static feature encoding weights $(\mathbf{W}_s)$ that are trained, thus: $\mathbf{v}^t = \mathbf{W}_s  \mathbf{u}^t $ where $\mathbf{W}_s \in \mathbb{R}^{r\times D}$.

In order to sequentially encode time-lapsed features from single frames into one dimensional video visual features, a sequential encoder is used. In contrast to current methods such as~\cite{venugopalan2014translating} and~\cite{Yao2015} that adopt simple mean average encoders, in our framework we use an LSTM-based bi-directional encoder that enables us to learn attention weights with better sequential dependency. 
We define our encoder as  $\mathbf{h}^t =\mathcal{E}(\mathbf{v}^{t},\mathbf{h}^{t-1})$ where $\overrightarrow{\mathcal{E}}(\cdot)$ and $\overleftarrow{\mathcal{E}}(\cdot)$ are a forward and backward-direction LSTM recurrent functions respectively:
\begin{equation}
\begin{aligned}
\overrightarrow{\mathbf{h}}^t &= \mathbf{\overrightarrow{\mathcal{E}}}({\overrightarrow{\mathbf{W}}_e}\mathbf{v}^{t} + \overrightarrow{\mathbf{U}}_e\overrightarrow{\mathbf{h}}^{t-1})\mbox{,}\\
\overleftarrow{\mathbf{h}}^t &= \mathbf{\overleftarrow{\mathcal{E}}}(\overleftarrow{\mathbf{W}_e}\mathbf{v}^{t} + \overleftarrow{\mathbf{U}_e}\overleftarrow{\mathbf{h}}^{t+1})\mbox{,}
\label{eq:biencoder}
\end{aligned}
\end{equation}

where $\mathbf{W}_e$ are the encoder weights for frame features $\mathbf{v}$ and $\mathbf{U}_e$ are the encoder transition matrix between hidden states. Both, $\mathbf{W}_e$ and $\mathbf{U}_e$ correspond to the weights from all the gating functions of the LSTM encoder~\cite{hochreiter1997long}.

The combined feature encoding for both the bi-directional encoder is defined as: 
\begin{equation}
\begin{aligned}
\boldsymbol{\nu} &= [\overrightarrow{\mathbf{h}}^t,\overleftarrow{\mathbf{h}}^t, \mathbf{v}^t]\mbox{,}
\label{eq:biencoder}
\end{aligned}
\end{equation}
where $[...]$ indicates concatenation.

\subsection{Conditional Decoder and Soft Attention}
Similar to other recent approaches~\cite{Yao2015}, our decoder makes use of soft attention weights denoted as $\mathbf{C}$ in equation~\ref{eq:decoder}, which are learned during training. However, in contrast to~\cite{Yao2015}, our input to the soft attention weights are reduced through a bi-directional neural network encoder. We also use a regularized LSTM unit introduced in~\cite{zaremba2014recurrent}. Our decoder with soft attention then takes the following formulation:
\begin{equation}
\begin{aligned}
\mathbf{g}_i & =  \sigma(\mathbf{W}^{\star}\mathbf{x}_i+\mathbf{U}^{\star}\mathbf{h}_{i-1}+\mathbf{C}^{\star}\boldsymbol{\nu})\\
\mathbf{z}_{i} & =  g(\mathbf{W}^{z}\mathbf{x}_{i}+\mathbf{U}^{z}\mathbf{h}_{i-1}+\mathbf{C}^z\boldsymbol{\nu})\\
\mathbf{c}_{i} & = \mathbf{i}_{i}\odot \mathbf{z}_{i}+\mathbf{f}_{i}\odot \mathbf{c}_{i-1}\\
\mathbf{h}_{i} & =  \mathbf{o}_{i}\odot g(\mathbf{c}_{i})\mbox{,}
\end{aligned}
\label{eq:decoder}
\end{equation}
where $\sigma$ and $g$ represent a sigmoid and a hyperbolic tangent function respectively. The operation $\odot$ represents a Hadamard product. The input ($\mathrm{i}$), forget ($\mathrm{f})$ and output ($\mathrm{o}$) gating functions of the LSTM unit correspond to $\mathbf{g}_{t} =\{\mathbf{i}_{t}, \mathbf{f}_{t}, \mathbf{o}_{t}\}$ with its corresponding weights $\mathbf{W}^{\star}$ and $\mathbf{U}^{\star}$, where $\star=\{\mathrm{i},\mathrm{f},\mathrm{o}\}$.

\subsection{MTL Encoder}
\label{sec:dual-dec}
Multitask learning (MTL) is a paradigm that studies the problem of estimating multiple functions jointly by exploiting shared structures in order to improve generalization~\cite{caruana1998multitask}.

We can treat our encoder-decoder framework as a function $f$ to approximate through MTL. However, because the relation between their tasks is non-linear, then solving this MTLE problem is non-trivial~\cite{CilibertoRRP17}. Nonetheless, we consider the problem as a linear MTL approximation in order to solve it through a convex optimization.

Formally, given a set of functions:
$f_1, ...,f_n:\mathcal{X}\rightarrow\mathbb{S}$ and a corresponding set of training samples $(\mathbf{V}_l, \mathbf{X}_n)$, with $\mathbf{X}_i \in \mathbb{S}$ and $\mathbf{X}_i \in \mathbb{V}$, we define $\mathbb{S}$ as the semantic space where all the captions reside and $\mathbb{V}$ is the visual space with  $\mathbf{V}_l =\{\mathbf{v}^1, ...,\mathbf{v}^t\}, $where $l$ corresponds to the $l$-th video in the training data with its corresponding caption $\mathbf{X}_n$. In this sense, for each $f$ function, the following holds: $f_n(\mathbf{V}_l)=\mathbf{X}_n$.

We model task relations as a set of $\mathcal{P}$ functions with a constraint function $ \gamma :\mathbb{V}^l \rightarrow \mathbb{S}^{\mathcal{P}}$ and require $\gamma(f_1(\mathbf{V_l}),...,f_n(\mathbf{V_l}))=f'(\mathbf{V_l})$, where $f'(\mathbf{V}_l)$ corresponds to the ``true'' semantic position of $\mathbf{V}_l$ in $\mathcal{S}$. 

Our problem imposes a constraint in the range of $\gamma$, mainly, $\mathcal{X} \rightarrow \mathcal{C}$ to take values in the constraint set: 
\begin{equation}
\mathcal{C} = \big\{ \mathbf{y} \in \mathbb{S}^n \mid \gamma(\mathbf{y}) = f'(\mathbf{V}_l)\big\} \subseteq \mathbb{S}^n\mbox{,}
\end{equation}
Thus the goal is to find a good approximation $\hat{f}:\mathcal{X}\rightarrow \mathcal{C}$ for the following multi-task \textit{expected risk} minimization problem:
\begin{equation}
\begin{gathered}
\min_{f:\mathcal{X} \rightarrow \mathcal{C}} \mathcal{E}(f),\\
\mathcal{E}(f) = \frac{1}{N} \sum_{n=1}^N\frac{1}{M_n}\sum_{m=1}^{M_n}\mathcal{L}(f_n(\mathbf{x}_{mn}),\mathbf{y}_{mn})d\rho_n(\mathbf{x},\mathbf{y}),
\label{eq:multi_min}
\end{gathered}
\end{equation}
where $\mathcal{L}:\mathbb{R} \times \mathbb{R} \rightarrow \mathbb{R}$ is the loss function of prediction errors for each task $n=1,...,N$, $\rho_n$ is the distribution on $\mathcal{X} \times \mathbb{S}$ from where training points $(\mathrm{x}_{mn}, \mathrm{y}_{mn})_{n=1}^{m_n}$ have been sampled independently.
If $\mathcal{C}$ is a non-linear subset, the minimization in equation~\ref{eq:multi_min} is difficult to solve through convex optimization. Hence, we assume that $\mathcal{C}$ is a linear subset. Furthermore, similar to~\cite{dinuzzo2011learning}, we approximate equation \ref{eq:multi_min} by calculating a matrix of pair captions that resembles the positive semi-definite matrix $\mathcal{A}$ that encourages linear relations between the tasks discussed in~\cite{dinuzzo2011learning}. Furthermore, in contrast to ~\cite{dinuzzo2011learning}, we create and treat $\mathcal{A}$ as a sampling distribution rather than a second term to the solution in this way: 
\begin{equation}
\begin{gathered}
\min_{f=(f_1,...,f_n) \in \mathcal{H}^n} \lambda\sum_{n=1}^N\sum_{m=1}^M\mathcal{L}(f(\mathbf{x}_{m}),\mathbf{y}_{m}) \cdot \mathcal{A}_{ns}\langle n,c \rangle,
\label{eq:min_lin}
\end{gathered}
\end{equation}
where $\lambda$ is a normalization term, $\mathcal{L}$ is the loss function of the system, $\mathcal{A}= (\mathcal{A}_{ns})_{c,n=1}^N$ is called the Semantic Distance Matrix and $c$ is the index for the caption with the largest dissimilarity to the $n$ caption both discussed in~\ref{sec:SDM}. $\mathcal{H}$ is a reproducing kernel Hilbert space and in order to evaluate the minimization as a linear combination, we assume that the functions $f_n$ are part of $\mathcal{H}$.

The total cost is obtained by adding the total loss $\mathcal{L}$ of each $n$ independent task of the system along with a regularization term. Although, our method allows for any number of tasks, for practical purposes, we set $n=2$ in our model, and our loss function is defined as follows: 
\begin{equation}
\begin{multlined}
\sum_{n}\mathcal{L}(f(\mathbf{x}),\mathbf{y})\cdot \mathcal{A}\langle f_1,f_c \rangle =  \sum_i{-\log P^1_{\mathbf{\hat{x}}}} +\\ \sum_{i}{-\log P^c_{\mathbf{\hat{x}}}} 
 + \eta \sum_i{| P^1_{\mathbf{\hat{x}}}- P^c_{\mathbf{\hat{x}}}|}\mbox{.}
\label{eq:loss_reduced}
\end{multlined}
\end{equation}
where $f_1$ represents the main reference task and $f_c$ is the complement task to $f_1$. We explain the details on how $f_c$ is chosen in section~\ref{sec:multisingle}. $\eta$ is a parameter set to [0,1] depending if the videos belong to a single or multi-caption dataset respectively.
$P^n_{\mathbf{\hat{x}}}$ represents the cost of a probability centroid caption in semantic space related to task $n$. Concretely, given the probability prediction defined in equation~\ref{eq:inference} for all $n$ tasks, we obtain the centroid probability in the following way:
\begin{equation}
P_\mathbf{\hat{x}}=\sum_n{\frac{P^n_{\mathbf{x}}}{n}}=\frac{P^1_\mathbf{x}+...+ P^n_\mathbf{x}}{n} 
\end{equation}

\begin{figure}
  \centering
  \includegraphics[width=.4\textwidth]{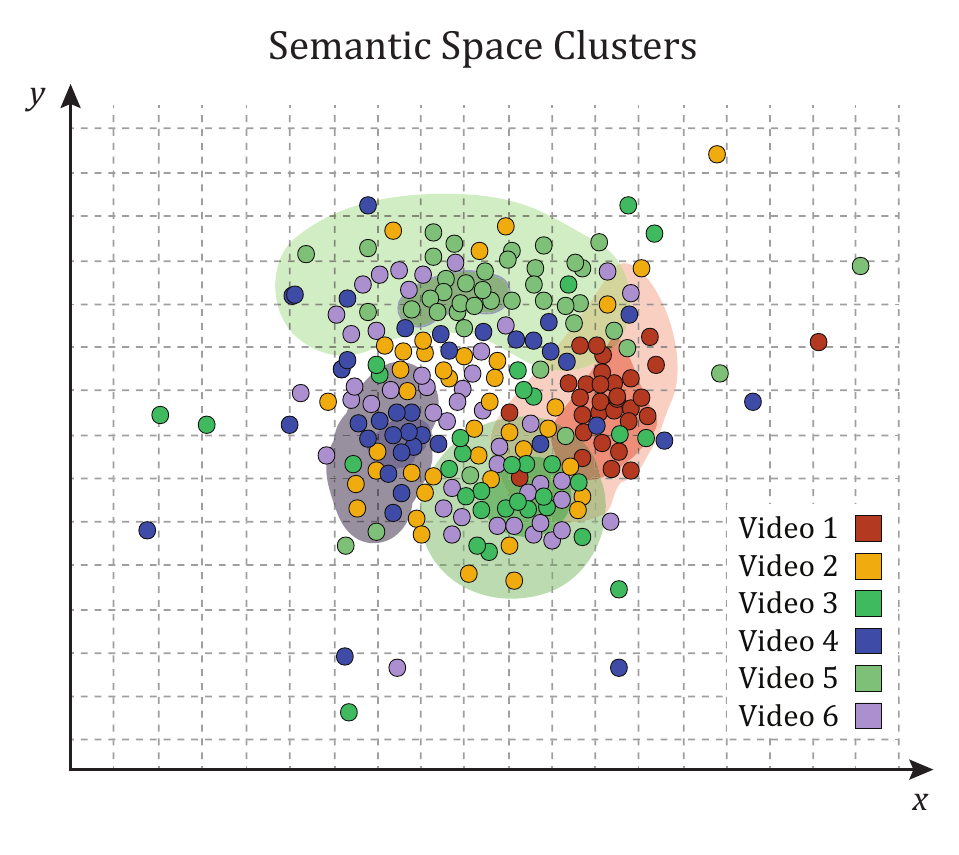}	
  \caption{Semantic space representation with t-SNE embedding of skip-thought vectors of captions in different videos. Each dot in the graph represents a caption of a video represented by a particular color.}
  \label{fig:skipspace}
\end{figure}

The idea of finding a centroid that represents all captions in a video comes from the observation that all captions that belong to a specific video lie near to each other when projected into a semantic space. We can build this semantic space by obtaining skip-thought~\cite{kiros2015skip}  vectors for each of the captions and use their projected space as their semantic space. If we were to visualize the t-SNE~\cite{maaten2008visualizing} reduction of such vectors as shown in figure~\ref{fig:skipspace}, we would observe that some of the captions that correspond to their respective videos form natural clusters. Our goal is then, to represent these clusters by their centroids which at a higher level represent the broad meaning of all the caption sentences.   

\subsubsection{Multiple vs. Single Caption}
\label{sec:multisingle}
The complement task $f_c$ introduced in equation~\ref{eq:loss_reduced} is obtained depending on the number of captions per video found in the dataset. In the case of multiple captions per video $f_c$ is chosen using a semantic distance matrix explained in section~\ref{sec:SDM}. If single caption samples are prevalent in the dataset, we augment the dataset with one additional caption per video that is obtained in the following way: Given a caption $\mathbf{X}= \{\mathbf{x}_1, ..., \mathbf{x}_i \}$, we obtain a function that will create a second caption $\mathbf{Y}$ based on the following rule: $\mathbf{Y} = \{\forall i, \mathbf{x}_i \in \mathbf{X}| \mathbf{x}_i \notin \mathbf{P}  \}$
where $\mathbf{P}$ is a set of stop words from a given dictionary. Thus, the sampled caption $\mathbf{Y}$ will vary from the original caption $\mathbf{X}$, where $\mathbf{Y}$ contains only keywords that represent objects (nouns) or actions (verbs) present in the video that we want the encoder and attention weights to emphasize without much consideration of the  grammatical structure of the resulting caption. This allows us to create another caption that is still somewhat different to the original one but with injected bias towards more semantically important words.

\subsubsection{Semantic Distance Matrix}
\label{sec:SDM}

Matrix $\mathcal{A}$ could be simply created from all the captions for one video sampled in any order. However, such simplistic approach would be no different to utilizing a single task without leveraging much the diversity of the many annotations per video and the advantages of a multitask approach.

Another approach would be to use all $n$ captions provided in the dataset for a particular video to initialize $n$ different tasks and decoders. However, such approach would increase linearly the number of parameters in the network and could be overwhelming to train. 

In order to take advantage of the gamut of information inherent in the ground truth annotations while being conservative in the growth of the parameters of our network, we use two decoders and their semantic properties of the corresponding captions to construct $\mathcal{A}$ in relation to how different or apart such decoders are in their semantic space.

More formally, let us define $\mathbf{\Lambda}$ as the set of captions that belong to one video and assume that there is at least one complementary caption $\mathbf{Y} $ such that, $\mathbf{\Lambda}=\{ \forall~\mathbf{X} ~ \exists~  (\mathbf{X}, \mathbf{Y}) : \mathbf{Y} = \Delta(\mathbf{X}) \land \mathbf{Y} \in \mathbf{\Lambda}\}$, where $\Delta$ is a semantic distance function and $|\mathbf{\Lambda}| = n$.

In order to compute $\Delta$, we first obtain skip-thought vectors~\cite{kiros2015skip} $\mathcal{S}$   from each caption video which will help us project our raw captions into a semantic space where their semantic distance can be compared.  Thus, given captions $\mathbf{X}$ and $\mathbf{Y}$ we obtain their respective skip-thought vectors: $\mathbf{u} = \mathcal{S}(\mathbf{X})$ and  $\mathbf{v} = \mathcal{S}(\mathbf{Y})$. The semantic function $\Delta$ is then obtained as follows : 

\begin{equation}
\begin{aligned}
\Delta(\mathbf{u},\mathbf{v}) &= 1 - \frac{\mathbf{u} \cdot   \mathbf{v} }{\lVert \mathbf{u} \rVert_2 \lVert \mathbf{v} \rVert_2}\mbox{,}\\
\mathcal{A}_{nc}&= {\arg \max}_{c} \Delta(\mathcal{S}(f_n),\mathcal{S}(f_c))
\end{aligned}
\end{equation}
where $\Delta$ is a measurement between 0 and 1 that indicates the dissimilarity between $\mathbf{u}$ and $\mathbf{v}$. Our semantic distance matrix allow us to leverage the semantic diversity embedded in all the different captions that belong to a particular video and thus represent better their respective semantic cluster.

\subsubsection{Inference}
During inference, the probability distribution of obtaining a predicted caption description $\mathbf{X}$ for a video $\mathbf{F}$ given network parameters $\boldsymbol{\Theta}$ of a task $n$ is given by:
\begin{equation}
\begin{aligned}
P(\mathbf{X} \mid \mathbf{F}, \boldsymbol{\Theta}_{n}) = \prod_{i=1}^I P(\mathbf{x}_i \mid \mathbf{x}_1,...,\mathbf{x}_{i-1},\boldsymbol{\nu},\boldsymbol{\Theta}_{n})
\end{aligned}
\end{equation}
Because in our model we train more than one independent task, the question of which decoder should we use during inference could arise. Here we refer to equation~\ref{eq:loss_reduced} where the third term in the equation acts as a regularization term and aims to reduce the distance of the decoders during training and forcing both decoders to produce similar results. For this reason, during inference, we fix the weights to any of the decoders in task $n$ assuming that any of them would produce similar results to the other. The probability $P$ of a word caption $\mathbf{x}$ for a task $n$ is given as follows:

\begin{equation}
\begin{aligned}
P^n_\mathbf{x}&=P(\mathbf{x}_i|\mathbf{x}_{<i},\boldsymbol{\nu},\boldsymbol{\Theta}_{n}),\\
 P(\mathbf{x}_i|\mathbf{x}_{<i}, \boldsymbol{\Theta}_{n}) &= \textnormal{softmax}(\mathbf{W}_{d} \mathbf{h}_i)
\end{aligned}
\label{eq:inference}
\end{equation}
where $\boldsymbol{\Theta}_{n}$ represent the network parameters for decoder $n$ at inference. $\mathbf{W}_d$ is a weight matrix that maps the decoder`s hidden state to a distribution over the vocabulary $V$.

\section{Results}
In this section we present several comparisons of our method with other current and state of the art methods for video and movie description.  We divide this section into multi-caption and single-caption datasets.

\subsection{Multi-caption Datasets}

\setlength{\tabcolsep}{2pt} 
\begin{table*}[!ht]
\caption{Multi--Caption Datasets}
\begin{center}
\begin{tabular}{lcccccc}
\hline\noalign{\smallskip}
Dataset & Videos & Clips & Sentences & Domain & Sentence Source & Description  \\
\noalign{\smallskip}
\hline
\noalign{\smallskip}
\noalign{\smallskip}

 MSVD~\cite{msvd_ds} & 2,089 & 2,089 & 85,550 & Open & Crowd & Crowd-sourced captions of short Youtube videos.  \\
 MSR-VTT~\cite{msrvtt_ds} & 10,000 & 10,000 & 200,000 & Open & Crowd & Crowd-sourced captions of short Youtube videos. \\
 TRECVID-VTT~\cite{trecvid} & 1,880 & 1,880 & 3,760 & Open & Crowd & Crowd-sourced captions of short Vine videos.  \\
 
\noalign{\smallskip}
\hline

\end{tabular}
\label{table:multi-caption}
\end{center}
\end{table*}
\setlength{\tabcolsep}{1.4pt}

In this subsection, we describe our results in the context of two popular and well known datasets for video description in the wild, mainly, MSVD~\cite{msvd_ds} and MSR-VTT~\cite{msrvtt_ds}. We also present our results on a somewhat newer dataset for video description which was part of the TRECVID challenge 2017.

\setlength{\tabcolsep}{4pt}
\begin{table}
\begin{center}
\caption{MSVD Dataset}
\begin{tabular}{lcccc}
\hline\noalign{\smallskip}
Model & BLEU & METEOR \\
\noalign{\smallskip}
\hline
\noalign{\smallskip}
 \noalign{\smallskip}
 FGM~\cite{thomason:coling14} & 0.137 & 0.239  \\
  DR-LSTM~\cite{venugopalan2014translating} & 0.312 & 0.269  \\
 Yao ~\cite{Yao2015} & 0.419 & 0.296   \\
 S2VT MT~\cite{venugopalan2016improving}& 0.421& 0.314 \\
 HRNE~\cite{pan2016hierarchical} & 0.438&  0.331 \\
  h-RNN-VGG~\cite{yu2016video} & 0.499& 0.326   \\
 SCN~\cite{SCN_CVPR2017} & 0.511&  \textbf{0.335}  \\
 \noalign{\smallskip}
 \hline
 \noalign{\smallskip}
Baseline + C3D &0.411& 0.286 \\
Baseline + GoogleNet  & 0.455 &0.304  \\
Baseline + ResNet  & 0.493 & 0.320  \\
MTLE + GoogleNet & 0.497 & 0.319  \\
\noalign{\smallskip}
\textbf{MTLE + ResNet} & \textbf{0.530} & {0.318}  \\
\noalign{\smallskip}
\hline
\end{tabular}
\label{table:msvd}
\end{center}
\end{table}

\textbf{MSVD}: 

The Microsoft Research Video Description Corpus (MSVD) is one of the most widely known dataset in the domain of video description. The dataset contains 1,970 open-domain video clips and over 85k English description sentences~\cite{msvd_ds}. As follow by other methods such as~\cite{Yao2015}, we split the dataset into training, validation and tests sets with a number of 1200, 100 and 670 video clips respectively.

Table~\ref{table:msvd} shows a comparison of quantitative results of our method with other methods published in recent years such as FGM~\cite{thomason:coling14}, DR~\cite{venugopalan2014translating}, Yao~\cite{Yao2015}, S2VT MT~\cite{venugopalan2016improving}, h-RNN-VGG~\cite{yu2016video},
 HRNE~\cite{pan2016hierarchical} and SCN~\cite{SCN_CVPR2017}. These methods were published in the last couple years and are the state of the art in video description in the MSVD dataset.
In this table we also present our own implementation of~\cite{Yao2015}, a soft attention method which we name ``Baseline'' throughout this and other tables. We experimented with different CNN features such as GoogleNet, ResNet and C3D. Overall, ResNet showed the best performance on this and other datasets while C3D performed the worst. 
In MSVD our MTLE method shows competitive results in the Meteor metric and significant improvements on Bleu, with the best performance advantage over other methods in the latter one, even when using only static ResNet features. Notice also that even when using shallower networks for feature extraction such as GoogleNet, our model MTLE still shows improvement over the baseline.

\textbf{MSR-VTT}:
The Microsoft Research Video to Text (MSR-VTT) dataset is a recently released dataset that was part of a yearly ACMM grand challenge starting in 2016 and concluding in 2017~\cite{msrvtt_ds}. The dataset contains 10k open-domain video clips that are described by 200k crowd-sourced sentences. With a total video length of approximately 41 hours, the MSR-VTT dataset is the largest public multi-caption dataset, and consists of two versons. The 2016 version was used in the ACMM'16 grand challenge and contains 10k total videos split into training, validation, and test set by the challenge organizers. The 2017 version was used in the ACMM'17 grand challenge, and contains 13k total videos, with the original 10k videos from the previous challenge used as the training split, and an additional 3k videos used as the test set. In this paper, we analyze the results of our experiments performed on MSR-VTT 2016 with corresponding splits provided by the authors, in addition to a $K$-fold cross-validation with $k=10$.

\setlength{\tabcolsep}{4pt}
\begin{table}[ht!]
\caption{MSR-VTT 2016 Dataset}
\begin{center}
\begin{tabular}{lcccc}
\hline\noalign{\smallskip}
  Model &BLEU   &METEOR  & ROUGE  &  CIDEr \\
\noalign{\smallskip}
\hline
\noalign{\smallskip}

\noalign{\smallskip}

Xu (2016)~\cite{Xu2016,vtt-supmat} & 0.366 & 0.259 & - & -\\

\noalign{\smallskip}
\hline
\noalign{\smallskip}

Base. Googlenet & 0.377 & 0.249 & 0.576 & 0.391\\
Base. Resnet & 0.386 & 0.263 & 0.591 & \textbf{0.426}\\ 
MTLE Googlenet & 0.378 & 0.256 & 0.581 & 0.396\\
MTLE Resnet & \textbf{0.392} & \textbf{0.266} & \textbf{0.593} & 0.421\\ 

\noalign{\smallskip}
\hline
\medskip
\end{tabular}
\label{table:vtt16}
\end{center}
\end{table}
\setlength{\tabcolsep}{1.4pt}

\textbf{TRECVID-VTT}:
The TRECVID 2016 Video to Text (TRECVID-VTT) dataset contains over 50k Twitter Vine videos and was released initially as part of a TRECVID competition. 1880 of the videos were labeled with two captions, each provided by a different human annotator~\cite{trecvid}. Table~\ref{table:trecvid} shows a comparison of our method with our baseline with GoogleNet and ResNet features. Notice that our MTLE method significantly outperforms both baselines with their respective features.

\setlength{\tabcolsep}{4pt}
\begin{table}[!ht]
\begin{center}
\caption{TRECVID 2016 VTT Dataset}
\label{table:trecvid}
\begin{tabular}{lcccc}
\hline
   &BLEU   &METEOR  &  CIDEr   \\

\hline

  Baseline + GoogleNet & 0.1010 & 0.1445 &  0.3282 \\
    Baseline + ResNet & 0.1138 & 0.1477 &  0.3819 \\
    MTLE + GoogleNet & 0.1104 & 0.1477 &  0.3740 \\
 {\bf MTLE + ResNet} & \textbf{0.1217} & \textbf{0.3737} &  \textbf{0.4230} \\

\hline
\end{tabular}
\end{center}
\end{table}

\subsection{Single-caption Datasets}

\setlength{\tabcolsep}{2pt} 
\begin{table*}[!ht]
\caption{Single--Caption Datasets}
\begin{center}
\begin{tabular}{lcccccc}
\hline\noalign{\smallskip}
Dataset & Videos & Clips & Domain & Sentence Source & Description & Total Length (h) \\
\noalign{\smallskip}
\hline
\noalign{\smallskip}
\noalign{\smallskip}

 TACoS~\cite{tacos:regnerietal:tacl} & 127 & 18,227 & Cooking & Crowd & Actions and people in cooking videos. & - \\
 M-VAD~\cite{mvad_ds} & 92 & 48,986 & Movie & Professional & Movie description service annotations. & 84.6 \\
 MPII-MD~\cite{mpii_ds} & 94 & 68,337 & Movie & Professional &Movie description service annotations. & 77.8 \\
 LSMDC~\cite{lsmdc2017} & 200 & 128,085 & Movie & Professional & Movie description service annotations. & 147.0 \\
 
\noalign{\smallskip}
\hline

\end{tabular}
\label{table:single-caption}
\end{center}
\end{table*}
\setlength{\tabcolsep}{1.4pt}

The most relevant and well-known single-caption datasets for movie description research are M-VAD~\cite{mvad_ds}, MPII-MD~\cite{mpii_ds} which later were combined into one dataset called LSMDC~\cite{lsmdc_ds}. Here we give a brief overview of these datasets.

\textbf{M-VAD}:
The Montreal Video Annotation Dataset (M-VAD) consists of nearly 49k movie clips from 92 movies. The movies are paired with over 55k sentences transcribed from the Descriptive Video Service (DVS) narration of each movie~\cite{mvad_ds}. The average length of movie clips is 6.2 seconds, spanning a total runtime length of approximately 84 hours. 

\textbf{MPII-MD}:
The Max Planck Institute for Informatics Movie Description (MPII-MD) dataset provides over 68k movie clips and sentences from 94 movies. Sentences are derived from both the Descriptive Video Service (DVS) narration of each movie, and the movie's written script~\cite{mpii_ds}. The average length of movie clips is 3.9 seconds, with a total runtime length of approximately 73 hours. 

\textbf{LSMDC}:
The Large Scale Movie Description Challenge (LSMDC) dataset combines the videos and captions from the M-VAD and MPII-MD datasets, and have been used during the last years as part of an annual automatic movie description challenge~\cite{lsmdc_ds}. LSMDC is one of the largest datasets for movie description totaling 200 movie videos and 128,085 movie clips and 128,118 sentences.  with  Because LSMDC is a comprehensive combination of M-VAD and MPII-MD and more challenging dataset, our results for this paper are focused on the LSMDC dataset.

LSDMC differs from the other datasets such as TRECVID-VTT and MSR-VTT in that captions were not crowd-sourced, but instead taken from a professional movie description service. This offers a higher quality in the annotations, however, the numbers of sentences per video is limited to one which makes it a yet more challenging captioning problem because of the lack of training data with regards to the descriptions. Table~\ref{table:uni-multi} shows a comparison of our method with our baseline. For these results we use all the splits provided with the dataset. Also, notice that the results in Table~\ref{table:uni-multi} show better performance than the results from the competition in Table~\ref{table:lsmdc}. This is due to a better hyper-parameter tuning  in the latest version of our code.

\setlength{\tabcolsep}{4pt}
\begin{table}[!ht]
\begin{center}
\caption{LSMDC (M-VAD + MPII-MD) Dataset}
\label{table:uni-multi}
\begin{tabular}{lcccc}
\hline
   &BLEU   &METEOR  &  CIDEr  \\
\hline
    Baseline + ResNet & 0.004 &0.052 & 0.078  \\
  MTLE + GoogleNet & 0.004& 0.054 &0.074  \\  
 {\bf MTLE + ResNet } & \textbf{0.005} & \textbf{0.055} & \textbf{0.087} \\
 
\hline

\end{tabular}
\end{center}
\end{table}
\setlength{\tabcolsep}{1.4pt}

\subsection{Human Evaluation}
It has been noted that common metrics used to evaluate captioning systems do not resemble the performance of real human evaluators~\cite{vedantam2015cider}. Although traditional metrics such as BLEU, METEOR, among others offer a broad, quantitative comparison of ground truth sentences to predicted sentences, these metrics do not capture the semantics that humans would otherwise understand. We can see this conundrum in the numbers of the metrics reported in the LSMDC competition shown in Table~\ref{table:lsmdc} where some of the methods with relative lower metric numbers were ranked higher by human evaluators as being more useful to the blind as shown in Table~\ref{table:humaneval}.  

Evaluating metric scores of the predictions with their respective ground truth is not trivial. Ranking methods based on metric numbers differ from that of human evaluation ranking mainly because of the difficulty of the task. The abstraction of this task of rating comparing semantics of words in the English language is difficult to quantify numerically.  

Due to the limitations of current metrics, a human evaluation was performed among the competing methods at LSMDC 2017.
The evaluation consisted of randomly choosing 1000 video clips from LSMDC and the corresponding predictions from each of the competing methods. The predictions were provided to three independent human evaluators who ranked the predictions of each method on a scale of 1 through 5, where higher is better. The criteria given to the human evaluators was to rank the video captions based on how helpful they are to a blind person.

Table~\ref{table:humaneval} shows a comparison of the average scores received from the human evaluators during the LSMDC 2017 competition for each one of the top five methods~\cite{lsmdc2017}. ``Reference'' corresponds to the ground truth which was provided in the dataset by a DVS system. It is worth to note that the methods Fcrerank~\cite{kaufmantemporal} and PostProp~\cite{dong2016early} were the winners of LSMDC 2016 and ACMM grand chanllenge 2016 respectively.

\setlength{\tabcolsep}{4pt}
\begin{table}[!ht]
\begin{center}
\caption{LSMDC 2017 Competition}
\label{table:lsmdc}
\begin{tabular}{lcccc}
\hline\noalign{\smallskip}
Model & BLEU & METEOR  &  ROUGE & CIDEr  \\
\noalign{\smallskip}
\hline
\noalign{\smallskip}
Fcerank & 0.006 & 0.057 & 0.143 & 0.113\\
PostProp & 0.010 & 0.072 & 0.163 & 0.106\\
FuseNet & 0.005 & 0.055 & 0.142 & 0.083\\
Attn2l & 0.004 & 0.066 & 0.158 & 0.073\\
Moroni (\textbf{Ours}) & 0.003 & 0.052 & 0.134 & 0.073\\
PostProp-2 & 0.001 & 0.038 & 0.075 & 0.048\\
LSMDC 2016 & 0.006 & 0.058 &0.134 & 0.101\\
\noalign{\smallskip}
\hline
\noalign{\smallskip}

\end{tabular}
\end{center}
\end{table}
\setlength{\tabcolsep}{1.4pt}


\setlength{\tabcolsep}{4pt}
\begin{table}[!ht]
\begin{center}
\caption{Human Evaluation from LSMDC 2017}
\begin{tabular}{lcc}
\hline
   &Human Score    \\
\hline

  Reference (Human) & 4.46  \\
   {\bf MTLE (Ours) } &  \textbf{2.50} \\
    Fcrerank~\cite{kaufmantemporal} & 2.18  \\
    PostProp~\cite{dong2016early} & 2.17  \\
    FuseNet~\cite{lsmdc2017} & 1.96  \\
    attn2l~\cite{lsmdc2017} & 1.68 \\

\hline
\end{tabular}

\label{table:humaneval}
\end{center}
\end{table}


Notice that our algorithm received the highest score from human evaluators in the task of evaluating how useful the predicted captions are to the visually impaired.

\begin{figure}[h!]
  \centering
  \includegraphics[width=3.2in]{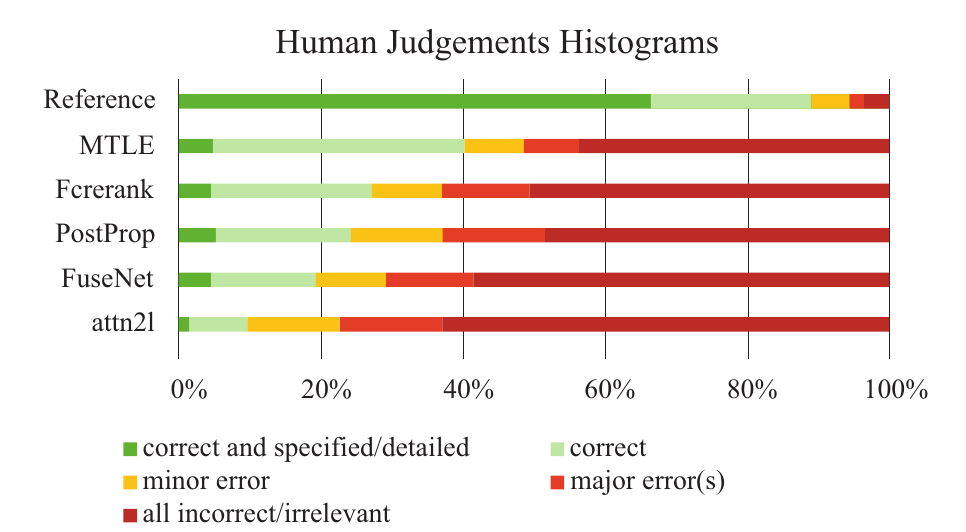}	
  \caption{Human judgment histograms of subjective scoring. Histograms indicate the percentage of errors classified as minor or major. Notice our method contains the smaller percentage of major and minor errors. }
  \label{fig:humaneval}
\end{figure}

In Figure~\ref{fig:humaneval}, we show a histogram of percentage of errors from our method and the top five other teams in the competition. Notice our method has the least percentage of minor and major errors.

\subsubsection{Qualitative Results}
\begin{figure}
  \centering
  \subfigure[MSVD]{\includegraphics[width=.47\textwidth]{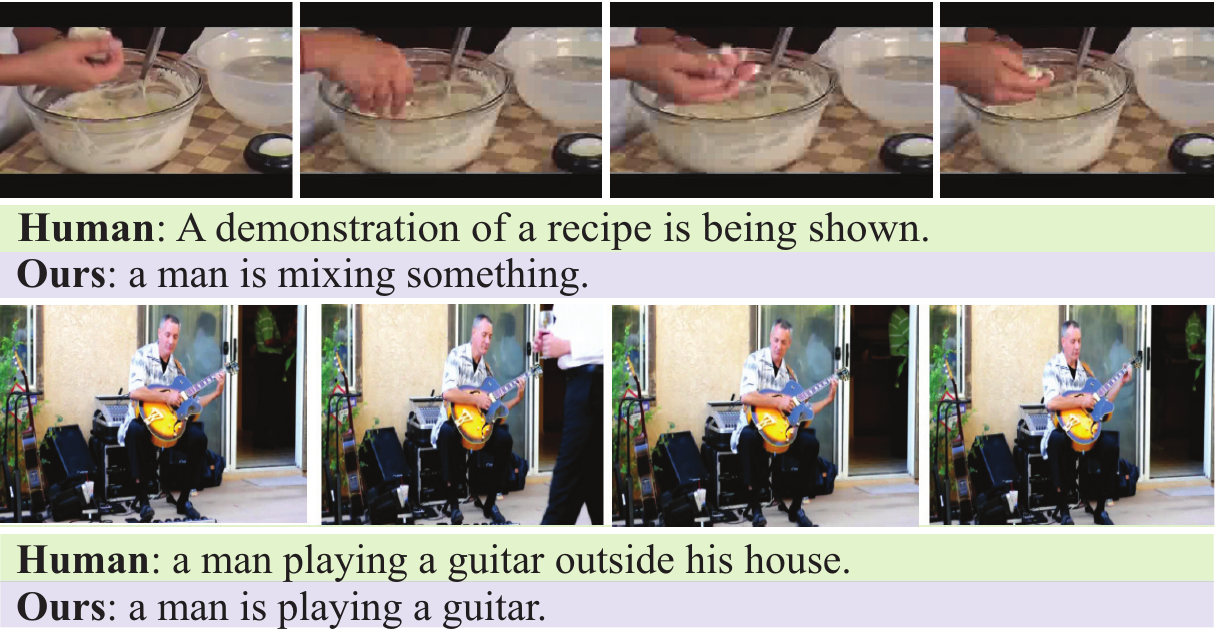}\label{fig:youtube}}\\
  \subfigure[MSR-VTT]{\includegraphics[width=.47\textwidth]{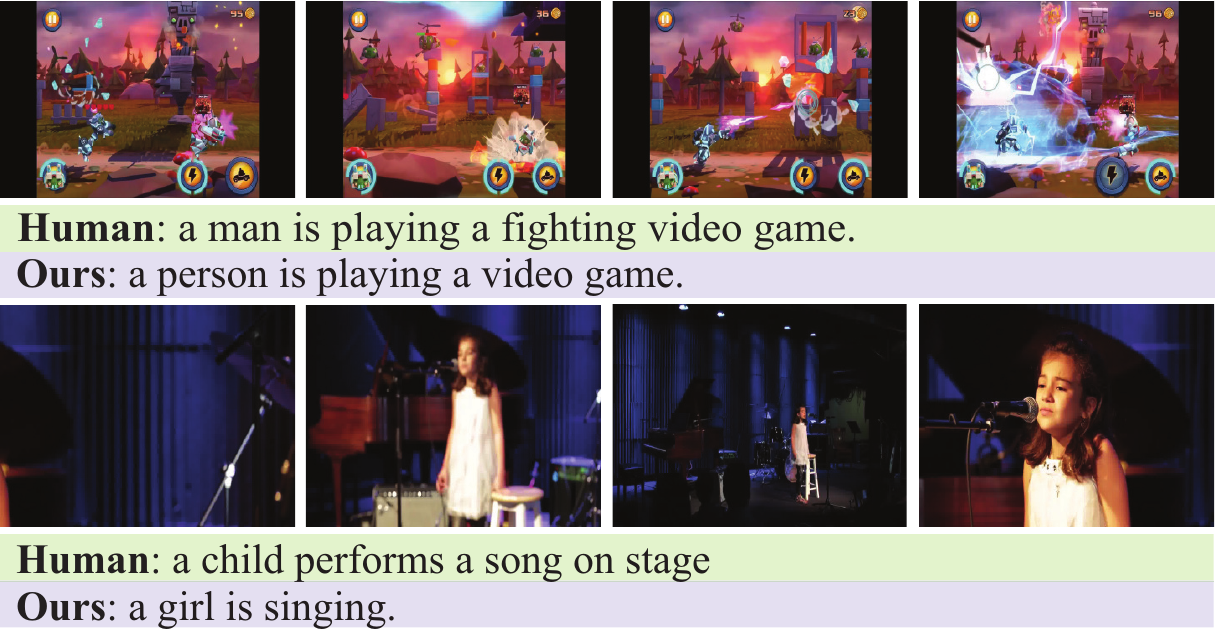}\label{fig:vtt}}
  \subfigure[LSMDC]{\includegraphics[width=.47\textwidth]{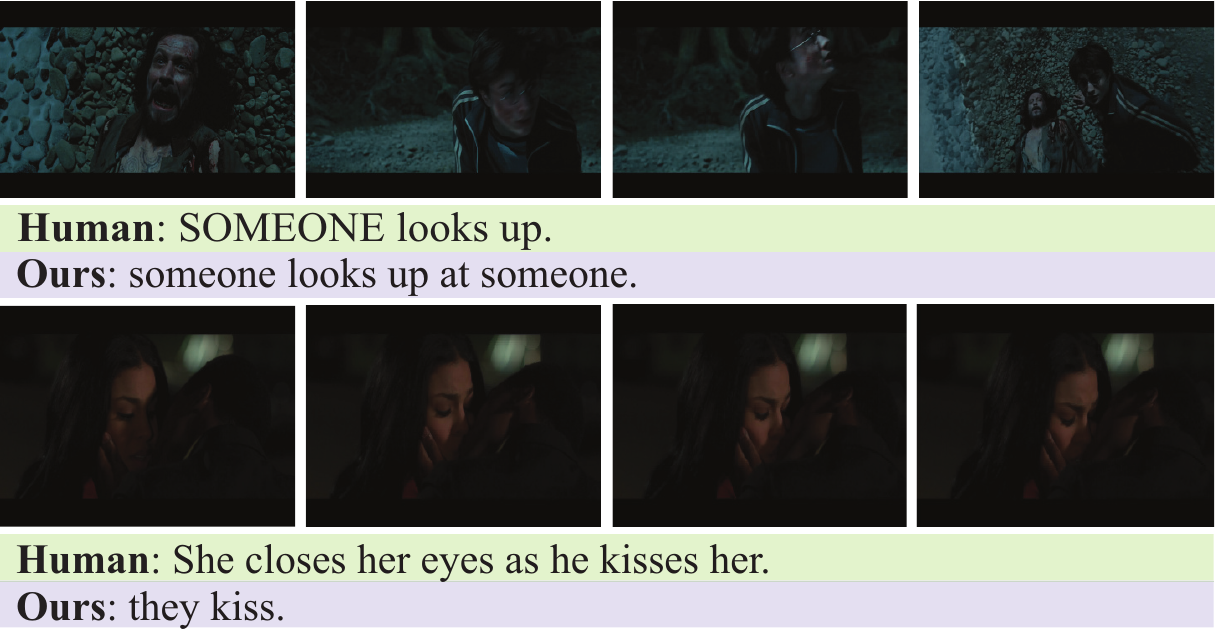}\label{fig:mpii}}
  \subfigure[TRECVID-VTT]
  {\includegraphics[width=.47\textwidth]{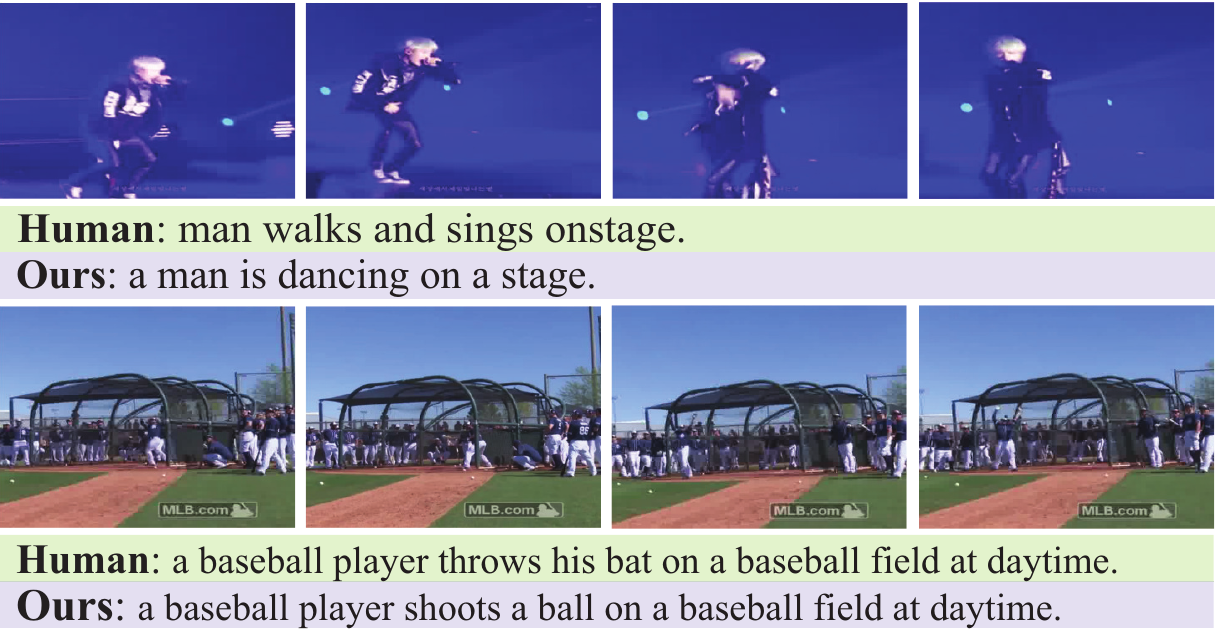}\label{fig:trecvid}}
  \caption{Qualitative results of our method on \ref{fig:youtube} MSVD, \ref{fig:vtt} MSR-VTT, \ref{fig:mpii} LSMDC, and \ref{fig:trecvid} TRECVID datasets.} 
  \label{fig:quaresults}
\end{figure}

Figure~\ref{fig:quaresults} shows qualitative results of our approach for uni/multi-label videos on the MSVD, MSR-VTT, LSMDC and TRECVID datasets. It is worth to notice that in some instances our system performs near human performance such as the second video of TRECVID~\ref{fig:trecvid} and the first video of LSMDC~\ref{fig:mpii}. In the second video of MSR-VTT~\ref{fig:vtt} our method even helps recognize the gender of the person, something the ground truth did not provide. Figure~\ref{fig:quaresults} shows some of our best results. We include in the supplementary material some fail cases and other comparisons and details that could not be included in this draft because of space constraints. 

\textbf{Source Code}: 
The code use for the competition was released at:

\url{https://github.com/OSUPCVLab/VideoToTextDNN}

\section{Conclusion}
In this paper we have presented a novel multitask encoder-decoder framework for semantic video and movie description. Our method helps improve a video feature encoder by leveraging the diversity of captions and a multitask framework to solve a multitask loss function through a convex optimization.
Our method shows promising results and in a human evaluation was ranked the highest and most useful among other methods for helping the visually impaired. 

\section{Acknowledgement}
We would like to thank Li Yao for his code and support with such.
Thanks to Anna Rohrbach and the organizers of the LSMDC 2017 competition for providing us with data results and figures from the competition for the publication of this paper. We would also like to thank insightful comments and discussions from Steven Rogers and Vincent Velten. 

This work was funded by the ASEE SMART program. The documentation has been approved for public release by the U.S. Air Force 88th Air Base Wing with PA approval number 88ABW-2016-4621.

\clearpage
{\small
\bibliographystyle{ieee}
\bibliography{egbib}
}
\clearpage
\textbf{UPDATE:} During the writing of this paper it was made aware to us about Pasunuru method~\cite{PasunuruB17} which also uses a multitask approach. Although our method differs greatly from Paunuru's, a further comparison with this method will be done on an upcoming version of our paper. 

\end{document}